\documentclass[acmsmall, authordraft=false, review=false, screen=true, timestamp=false]{acmart}
\usepackage{subfigure}
\usepackage{multirow}
\usepackage[ruled]{algorithm2e} 

\AtBeginDocument{%
	\providecommand\BibTeX{{%
			\normalfont B\kern-0.5em{\scshape i\kern-0.25em b}\kern-0.8em\TeX}}}
		
\copyrightyear{2020}
\acmYear{2020}
\acmMonth{6}
\acmDOI{10.1145/nnnnnnn.nnnnnnn}

\begin{document}
\title{Automatic Comic Generation with Stylistic Multi-page Layouts and Emotion-driven Text Balloon Generation}  
\author{Xin Yang}
\email{xinyang@dlut.edu.cn}
\orcid{1234-5678-9012-3456}
\author{Zongliang Ma}
\email{liangzongma1997@mail.dlut.edu.cn}
\author{Letian Yu}
\email{yuley3012@mail.dlut.edu.cn}

\affiliation{%
  \institution{ Dalian University of Technology}
  \department{Department of Computer Science}
  \streetaddress{2 Linggong Road}
  \city{Dalian}
  \state{Liaoning}
  \postcode{116024}
  \country{China}}

\author{Ying Cao}
\authornote{The corresponding author.}
\email{caoying59@gmail.com}
\affiliation{%
  \institution{City University of Hong Kong}
  \city{Hong Kong}
  \country{China}}

\author{Baocai Yin}
\email{ybc@dlut.edu.cn}
\author{Xiaopeng Wei}
\email{weixp@dlut.edu.cn}
\author{Qiang Zhang}
\email{zhangq@dlut.edu.cn}
\affiliation{%
	\institution{ Dalian University of Technology}
	\department{Department of Computer Science}
	\streetaddress{2 Linggong Road}
	\city{Dalian}
	\state{Liaoning}
	\postcode{116024}
	\country{China}
}

\author{Rynson~W.H.~Lau}
\email{rynson.lau@cityu.edu.hk}
\affiliation{%
	\institution{City University of Hong Kong}
	\city{Hong Kong}
	\country{China}}

\begin{abstract}
	In this paper, we propose a fully automatic system for generating comic books from videos without any human intervention. Given an input video along with its subtitles, our approach first extracts informative keyframes by analyzing the subtitles, and stylizes keyframes into comic-style images. Then, we propose a novel automatic multi-page layout framework, which can allocate the images across multiple pages and synthesize visually interesting layouts based on the rich semantics of the images (e.g., importance and inter-image relation). Finally, as opposed to using the same type of balloon as in previous works, we propose an emotion-aware balloon generation method to create different types of word balloons by analyzing the emotion of subtitles and audios. Our method is able to vary balloon shapes and word sizes in balloons in response to different emotions, leading to more enriched reading experience. Once the balloons are generated, they are placed adjacent to their corresponding speakers via speaker detection. Our results show that our method, without requiring any user inputs, can generate high-quality comic pages with visually rich layouts and balloons. Our user studies also demonstrate that users prefer our generated results over those by state-of-the-art comic generation systems.
\end{abstract}

%
%


%
%



\thanks{
This work was supported in part by the National Natural Science Foundation of China under Grant 91748104, Grant 61972067, Grant 61632006, Grant U1811463, Grant U1908214, Grant 61751203, in part by the National Key Research and Development Program of China under Grant 2018AAA0102003, Grant 2018YFC0910506, in part by the Innovation Technology Funding of Dalian (Project No. 2020JJ26GX036).
}
%

\maketitle

\renewcommand{\shortauthors}{X. Yang et al.}

\section{Introduction}

Manga is a popular artwork and information dissemination media, mainly due to its convenient reading form as well as excellent use of storytelling techniques (in drawing, paneling, layout, etc.) for enriched and immersive reading experience. However, traditional manga creation is time-consuming, and it requires professional skills and various content materials. Especially, it is very hard for non-professionals to produce their own manga books. Nowadays, movies or videos are common and popular everywhere with the development of Internet technology, especially the mobile videos uploaded to YouTube per day will take more than 82 years to watch for a man {\cite{Zhang2016Video}}. Such massive amount of videos can be  a useful resource for the manga content creation and may save content creation time for designers. Thus, how to convert videos or movies to manga books will be an interesting and significant job.
There have been some existing work \cite{Cao2012Automatic, Ryu2008CINETOON, Jing2015Content} proposed to edit videos in a manga-like way.  Wang \textit{$et\,al.$} \cite{Wang2012Movie2Comics} proposed an automatic schema for turning a movie clip to comics. Chu \textit{$et\,al.$} \cite{Chu2015Optimized} proposed a system to transform image sequence into a comics-based presentation in an optimized way. However, how to automatically convert videos to high-quality manga books with visual richness and excellent storytelling abilities is still a challenging and unsolved problem. The key difficulties are: 1) A fully end-to-end manga generation system to convert videos to manga books is desirable in a fast-forward prototype design, especially for non-professionals. However, existing works either require significant user inputs to give good results or are fully automatic but generate over-simplified results that lack visual richness and expressiveness. 2) The use of diverse word balloons in comics, whose shapes and font sizes vary according to contents, can enhance the expressiveness of a manga book and reinforce the readers' perception of feeling and emotional states of characters \cite{Kurlander1996Comic}. Exiting works generally select some pre-defined shapes for balloon generation. This would result in generated outputs failing to express story as faithfully as real comic pages. Thus, how to allocate abundant types of word balloon shape according to the emotion of different character dialogues and the corresponding audio is important. In order to generate multi-page manga books, panels should be assigned to different pages.  In \cite{Chu2015Optimized}, visual coherence and reading pace are main factors of their page allocation. Compared with their method, semantic relation of keyframes is considered in our system, which can help to allocate the keyframes with semantic relation on the same page.

Our framework solves the problem of allocating the selected keyframes across pages via a genetic algorithm and then organize the keyframes on each page manga-style layouts. To synthesize professional-looking manga-style layouts, we adopt a data-driven layout approach \cite{Cao2012Automatic} that learns layout styles from manga data. However, instead of relying upon users to specify inputs, we extend their approach by automatically extracting their inputs from the keyframes, including  region of interest, importance rank, inter-frame relation. This makes our layout framework fully automatic without the need for user inputs of any form. Finally, we propose a data-driven, emotion-aware balloon generation model, which can generate different balloon shapes and dynamically adjust font sizes based on the emotion of subtitles and audio. The generated balloons are then placed at the right places by detecting who is speaking and the location of the speaker’s mouth. Furthermore, in order to allow users to exert some degrees of control on generated results, we build a user-friendly interface to impose users' constraints towards building more personalized designs and fining tune the results. Our experiments show that the user interface can save users’ time, while providing more freedom of creation.

In summary, our contributions are:
\begin{itemize}
	\item We propose a fully automatic system to generate manga books from videos of arbitrary types (TV series, movies, cartoons). Our system does not need any manual inputs from users, and can generate high-quality manga pages with rich visual effects and expressive storytelling.
	\item We propose a multi-page layout framework for generating stylistically rich panel layouts across multiple pages jointly, based on rich semantics extracted from video frames automatically. 
	\item We propose an emotion-aware model for balloon generation. Our generated balloons can adapt to the emotion of contents measured by sentiment in subtitles and audios, which could enhance the expressiveness of final results.
\end{itemize}

\section{Related Work}
\subsection{Manga Generation}
There are already some related works on generating manga or comics. Ryu \textit{$et\,al.$} \cite{Ryu2008CINETOON} proposed a semi-automated system to generate black/white comic books from an input movie. Keyframes are extracted from a given movie manually. Then a "comic cut converter" based on Mean Shift segmentation and Bilateral Filter is used for stylization. After that, they perform background effect stylization by separating foreground from background and add some effect to the background. Finally, stylized font, speed line, word balloon, and stylized icon are placed to the image by users. However, a problem is that user intervention runs through its entire system, which requires a lot of manual efforts. Wang \textit{$et\,al.$} \cite{Wang2012Movie2Comics} proposed a complete approach to generate comics. Although they claimed that their system is automatic, it still needed a pre-prepared script file with speech contents and speaker identities. Further, they used regular grid-based layouts and a fixed balloon shape, which impairs the visual interest of generated results. Jing \textit{$et\,al.$}  \cite{Jing2015Content} described a content-aware approach for manga generation and generated layouts by maximizing information contained in a page. Unfortunately, their approach works only for conversational videos (videos with conversation of characters, such as TV series) which make it can't be applied to the videos without conversation, such as vlogs. And the shape of word balloon in their work is fixed which makes it boring for readers to read the manga book.  Chu \textit{$et\,al.$} \cite{Chu2015Optimized} designed a system to convert an image sequence to a comics-based presentation by explicitly solving multi-page panel allocation through a labeling optimization. Unfortunately, they only generated regular panel layouts with limited styles by simply matching content importance with pre-defined templates and also used a fixed balloon shape. In summary, despite progress made in generating comics from videos, all prior works either require additional user inputs, or use simplified representation or methods for layout and balloon generation, which cause results to lack visual variety and expressiveness. In contrast, our method is fully automatic and can generate visually rich, multi-page layouts and various balloon shapes that are adaptive to the emotion of speakers. 

\subsection{Keyframe Selection}
Keyframe selection is important to generate an interesting manga book. According to our research, there are many existing methods \cite{Wang2012Movie2Comics, Chu2015Optimized, Jing2015Content, Qu2013An} for keyframe selection. Wang \textit{$et\,al.$} \cite{Wang2012Movie2Comics} employed different strategies to extract descriptive keyframes after subshot detection and classification.  \cite{Jing2015Content} extracted speaker-key-frames based on speaker detection, and then extracted keyframes from changing scene.  Chu \textit{$et\,al.$} \cite{Chu2015Optimized} first extracted keyframes from subshots and then employed a keypoint-based approach to  find near-duplicate frames and eliminate redundant keyframes.  Qu \textit{$et\,al.$} \cite{Qu2013An} proposed a keyframe extraction method based on the similarity comparison between the undetected frame and the last shot on HSV color space. Different from their work, our keyframe selection method contains two stages: we first divide the video into two kinds of shots: dialogue-shots (shots with corresponding subtitles) and non-dialogue-shots (shots without corresponding subtitles) based on the information in subtitles. Then, we perform two different strategies for selecting dialogue-keyframes from dialogue-shots and non-dialogue-keyframes from non-dialogue-shots to ensure the fluency of the story and enrich the content of the manga book.

\subsection{Stylization}
In order to generate manga-style images, stylization is performed in our system. Previous works of stylization mainly take two kinds of stylization: black-white stylization and colored stylization. Ryu \textit{$et\,al.$} \cite{Ryu2008CINETOON} got black-white style images by mean Shift segmentation and Bilateral Filter are combined to get black-white style images. Wang \textit{$et\,al.$} \cite{Wang2012Movie2Comics} applied an image abstraction method to get colored style images by modifying the contrast of visually important features.  Both of the two kinds of stylization are used in the work of  \cite{Jing2015Content}. Winnemöller \textit{$et\,al.$} \cite{Winnem2012XDoG} proposed an advanced approach for stylization with extended Difference-of-Gaussians based on the work of  \cite{Holger2011XDoG}. 
{In recent years, there are some deep learning methods for stylization, (e.g., \cite{gatys2016image, johnson2016perceptual, gu2018arbitrary, Park_2019_CVPR}), however, the stylization module is not the core of our system. Since the approach of \cite{Winnem2012XDoG} is simple and requires no post-processing,  we employ it for stylization in our complex system, which we find  works well.}

\subsection{Panel Layout}
Panel layout is used to present all the panels in a manga style. Early researches, such as \cite{Ryu2008CINETOON}, only took grid layout which assigned images to rectangular grids.  \cite{Wang2012Movie2Comics} and  \cite{Toyoura2012Film} designed several layout templates and chose one of the templates as the layout according to the matching degree between input panels and the pre-defined templates. For example, \cite{Wang2012Movie2Comics} represented their template and generated panel list as two numerical sequences and computed the distance of the two numerical sequences to select the best template.  Jing \textit{$et\,al.$} \cite{Jing2015Content} determined the local structure based on video content and generated the initial layout of a comic page automatically, then they performed layout optimization to get the final panel layout. Cao \textit{$et\,al.$} \cite{Cao2012Automatic} proposed a data-driven method for panel layout. Firstly, they created an initial layout that best fit the input artworks and layout structure model, according to a generative probabilistic framework. Then, the layout and artwork geometries were jointly refined using an efficient optimization procedure, resulting in a professional-looking manga layout.

In our system, we choose the method proposed by  \cite{Cao2012Automatic} for panel layout since it provides various styles learned from different manga series. However, their method is semi-automatic, and needs users to provide ROI (region of interest), importance of each panel, relation of panels and the number of panels presented on each page. Thus, we extend their method by automating the estimation of the required inputs from video frames.

\subsection{Text Balloon Generation and Placement}
Balloon shape is important for manga creation. There are some tools for text balloon generation, such as "Balloonist''(Horlick, 2016), "Comic Book Creator'' (Planetwide Games, 2005) ,and "SuperLame!''(SuperLame, 2015). However, all of these tools require some manual efforts.  Preu \textit{$et\,al.$} \cite{Preu2007From} applied textual analysis on a screenplay and extracts information to create two types of balloons: speech balloon and noise balloon (balloon without dialogue and just express the emotion of background). Balloons placement is controlled by a layout algorithm to keep reading order and avoiding occluding speaker's face or exceeding panel boundary. The accuracy of their approach is acceptable, but requires an input screenplay which is not easy to obtain. Hong \textit{$et\,al.$} \cite{Hong2010Movie2Comics} proposed a script-face mapping method to detect who is speaking, and then place the pre-generated word balloon near the speaker and make the balloon tip point to the speaker. However, screenplays of the movies are not easy to get and the tip of balloon may not point to the speaker correctly. Wang \textit{$et\,al.$} \cite{Wang2012Movie2Comics} proposed an improved approach. They used speech recognition to replace script-face mapping for speaker recognition to eliminate the need for the screenplay. However, their simplified balloon generation  only uses one type of balloon which reduces the variety and expressiveness of generated manga books.  Sawada \textit{$et\,al.$} \cite{Sawada2013Film} chose oval balloons and used feature tracking for speaker detection and word balloon placement. Chu \textit{$et\,al.$} \cite{Chu2015Optimized} proposed an optimized method for balloon generation. Balloon size and balloon shape are two factors they took into consideration. However, they selected the shape of balloons only by comparing several words in a subtitle with pre-prepared words.

In this work, we consider multiple types of balloon shapes and change both balloon shape and font size in balloons by analyzing the emotion of subtitles and audios. This allows us to produce the results that better express the pacing of stories and the feeling of characters. Then, we perform the speaker detection in \cite{Sawada2013Film} for word balloon placement.


\section{Methodology}

\begin{figure*}
	\centering
	\includegraphics[width=5.50in]{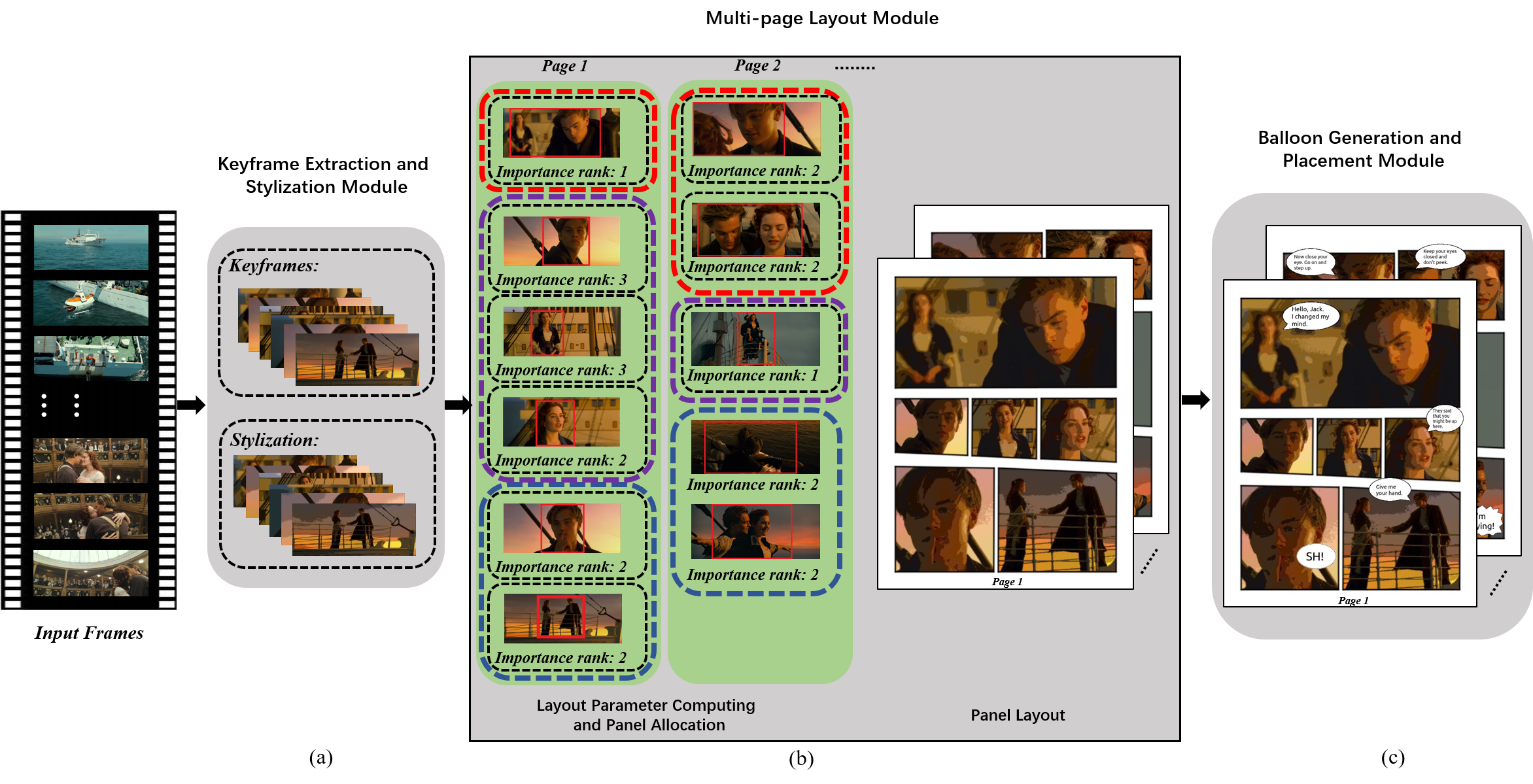}
	\caption[Fig.]{Overall pipeline of our system. (a): Keyframe Extraction and Stylization. (b): Automatic Multi-Page Layout Framework, red, purple ,and green dotted boxes mean different groups. (c): Balloon Generation and Placement. In step (a), we perform keyframe selection and stylization to get the stylized keyframes of the input video frames. In step (b), we first obtain four layout parameters of the frames including region of interest, importance rank, semantic relation, and allocate the frames across different pages. Then, we perform the layout algorithm in  \cite{Cao2012Automatic} for multi-page layout. In step (c), we designed an emotion-aware model for balloon generation and placement.}
	\label{fig_sim2}
\end{figure*}

Our key idea is to design our system in a fully automatic manner without any manually specified parameters or constraints. Meanwhile, we optionally introduce user interaction for more personalized design and diversities. As shown in Fig~\ref{fig_sim2}, the input of our system is a video with its subtitles file, and there are mainly three modules in our system, detailed as keyframe selection and stylization, multi-page layout generation, and balloon generation and placement. Firstly, representative keyframes are extracted and stylized from the input video in the keyframe processing module. Secondly, we perform an automatic multi-page layout framework to present all the panels in a manga-like way. Finally, we generate word balloons by analyzing the emotion of subtitles and audios, summarize multiple sentences, then perform lip motion detection to help for balloon placement. We will give more design and implementation details in the following sections.

\subsection{Keyframe Extraction and Stylization Module}
The input of our system is a video and its subtitles which contain dialogues and  corresponding start and end time information. We first select one frame every 0.5 seconds from the raw video. These selected frames can represent the raw movie. Then we use time information in subtitles and similarity between two consecutive frames to select informative keyframes. Finally, we perform stylization to convert ordinary images to manga-style images.

\begin{figure*}
	\centering
	\includegraphics[width=5.50in]{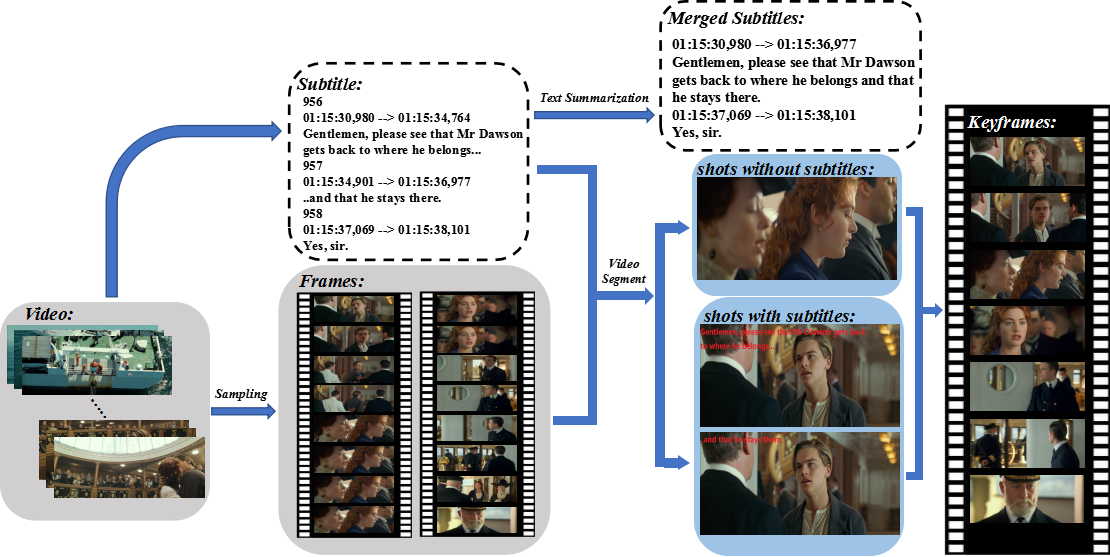}
	\caption{Pipeline of Keyframe Selection. Firstly, we select one frame every 0.5 second to get frames which can represent the input video. Meanwhile, the subtitles of the video is obtained from the video. Then, with the time duration of the subtitles, we divided the frames into two groups: frames with subtitles and frames without subtitles. For these two groups of frames, we take different strategies to pick up the representitive keyframes.}
	\label{fig_sim3}
\end{figure*}

\textbf{Keyframe Selection.} We make use of the time information for keyframe selection as shown in {Fig~\ref{fig_sim3}}. Firstly, we segment the video into shots using the start and end time of each subtitle. There are two kinds of shots: dialogue-shots (shots with subtitles) and non-dialogue-shots (shots without subtitles). For dialogue-shots, we compute the GIST \citep{Oliva2001Modeling} similarity between two consecutive frames obtained before because the GIST similarity is macroscopic and it is accurate for our work. If the foregoing GIST similarity is smaller, two frames differ more from each other. In our implementation, if the similarity is less than threshold $\theta$$_1$ we set, then the latter frame will be selected as a keyframe. If none of the frames corresponding to a subtitle are selected, we just pick the middle frame of the shot as a keyframe. Considering that more than one subtitle may correspond to a consecutive dialogue and the same scene, we compute the GIST similarity between consecutive keyframes obtained before. If the similarity is more than the threshold $\theta$$_2$ we set, we regard them as belonging to the same scene. Then, we will just keep one of them as keyframes and merge the subtitles. It is possible that we can select more than one keyframes during one subtitle, and we consider those keyframes to have semantic relations which will be used in multi-page layout. For non-dialogue-shots, we first select the frame most dissimilar to the frames in the shot. Then, in order to reduce redundancy, we compute the GIST similarity between the frame in this shot and the keyframes we selected before. Only if the similarity is less than we set before, this frame can be selected as a keyframe. Finally, the set of subtitles is then grouped by comparing the starting timestamp and keyframe's timestamp. Any subtitle starts within the time interval bounded by the start and end timestamp of a keyframe will be gathered together. In this way, we can extract informative keyframes and  get the semantic relevance of some keyframes which is important in multi-page layout.

\begin{figure}
	\centering
	\includegraphics[width=4.in]{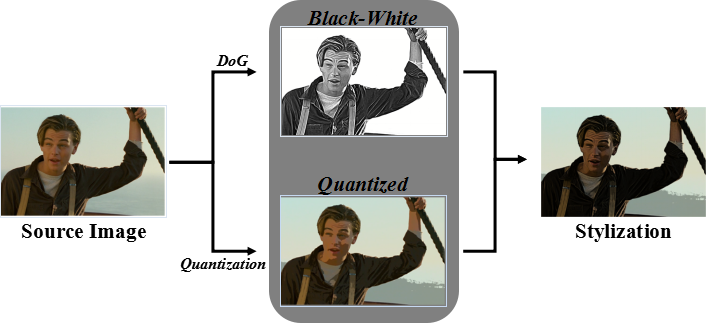}
	\caption{Pipeline of Stylization. Firstly, we convert the source image to black-white image via the extended DoG \cite{Winnem2012XDoG} and then get quantized image by color quantization. Finally, we combine these two kinds of image to get colored stylization.}
	\label{fig_sim4}
\end{figure} 

\textbf{Stylization.} As shown in { Fig~\ref{fig_sim4}}, we employ the extended Difference-of-Gaussians approach \cite{Winnem2012XDoG} to convert the source image to black-white image. For colored stylization, firstly, 128 level color quantization is executed to get the quantized image. Then, we get the DoG edges of the image using the algorithm in \cite{Winnem2012XDoG}. Finally, we get colored stylization by combining DoG edges and quantized image.

\subsection{Multi-page Layout Module}
Multi-page layout framework is used to automatically allocate and organize the panels across pages in visually rich layouts. In our framework, we first compute four key factors that are used to guide our multi-page layout generation, including region of interest of keyframes (ROI), importance rank of keyframes, semantic relation between keyframes, and the number of panels on one page. Then, We propose an optimization-based panel allocation method to assign the keyframes into a sequence of pages and use a data-driven manga-style layout synthesis approach to generate the layout for each of the pages.

\begin{figure}
	\centering
	\includegraphics[width=4.5in]{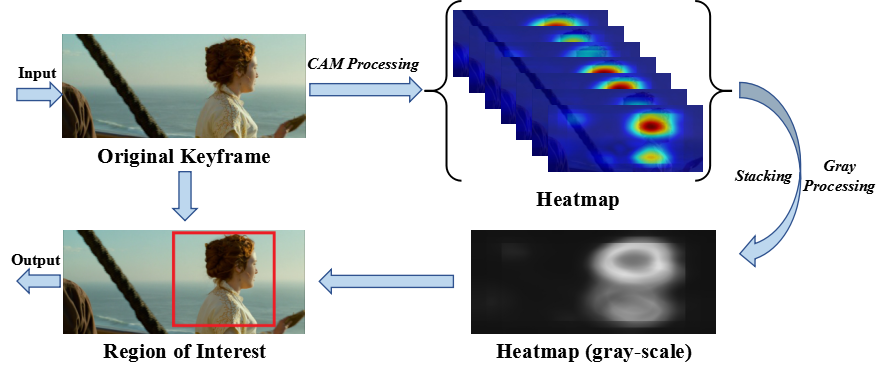}
	\caption{Processing of extracting the region of interest (ROI). We obtain different heatmaps by CAM processing from original keyframes. Then, we get gray-scale image of heatmap by stacking these heatmap and gray processing. Finally, we get ROI from gray-scale image of heatmap.}
	\label{fig_sim5}
\end{figure}

\begin{figure}
	\centering
	\includegraphics[width=5.5in]{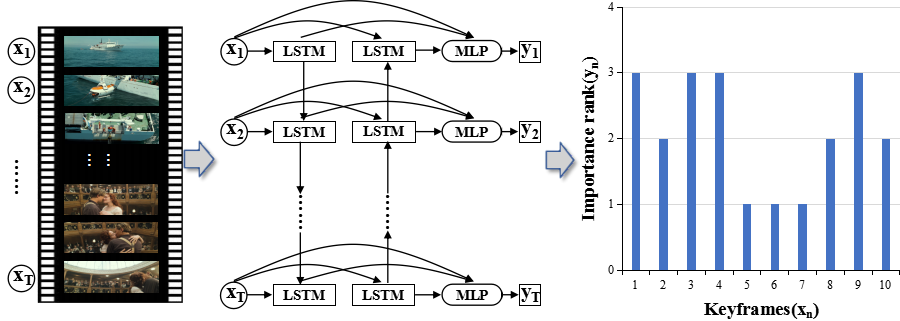}
	\caption{LSTM Model. The input \{x$_1$,x$_2$,...,x$_n$\} is the keyframes and the output \{y$_1$,y$_2$,...,y$_n$\} is the corresponding importance ranking of each keyframe.}
	\label{fig_sim6}
\end{figure}

\begin{figure}
	\centering
	\includegraphics[width=4.5in]{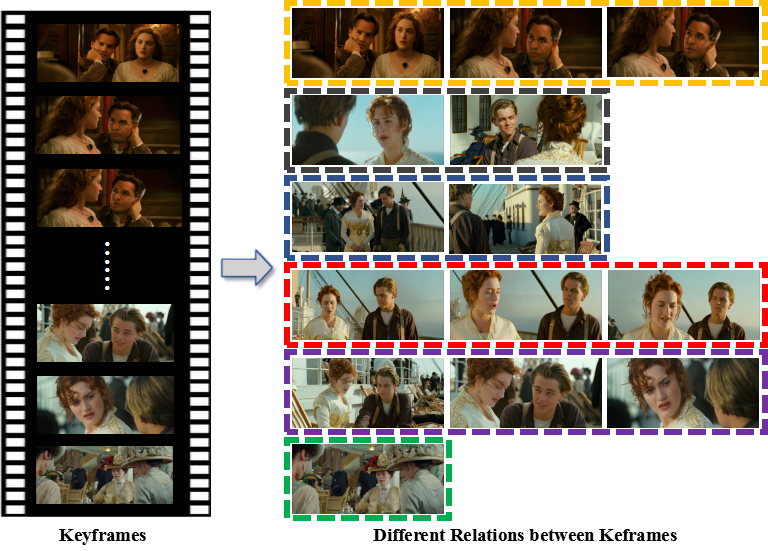}
	\caption{Different relation between keyframes. Frames in the same box have semantic relation while frames in different boxes do not have semantic relation.}
	\label{fig_sim7}
\end{figure}

\textbf{Layout Parameter Computing.} \begin{itemize}
	\item Region of Interest (ROI): We use a class activation mapping algorithm (CAM) for classification \cite{Zhou2016Learning} to find region of interest on a keyframe image because CAM can output a heatmap of an input image and the heatmap is usually the most attractive region of the input image. Specifically, an image is first input to a five cascaded layers convolutional neural network (CNN). The output features of the last CNN are the input to the global average pooling (GAP) layer. Then, a full connection layer is used to get the heatmap of the input image. As shown in Fig~\ref{fig_sim5}, in our implementation, we select the top seven scored heatmaps sorted by the classification score. We get the average of their activation values at the same position of these heatmaps, and convert them into a grayscale image. Finally, we compare the value of the grayscale image with the threshold $\theta$$_3$ we set. If the value of a point is greater than $\theta$$_3$, it is regarded to be within the ROI. And if the value of a point is less than $\theta$$_3$, it is regarded to be outside of the ROI. Then, a minimum bounding box is found by retaining all the within-points and reducing outside-points. The minimum bounding box is the ROI we find.
	{\item Importance Rank: In order to get the importance of each keyframe, we take use of the LSTM neural network used for video summarization \cite{Zhang2016Video}. As reported in \cite{Zhang2016Video}, the LSTM model can output frame-level importance scores representing the likelihoods of the frames being selected as a part of summary.
		The procedure can be summarized as follows: 1) As shown in Fig~\ref{fig_sim6}, we input a 1024-dimensions vector obtained by extracting the output of the penultimate later (pool 5) of the GoogleNet model \cite{Szegedy2015Going} to the LSTM model because it can modestly improve the performance over the same shallow features (i.e. color histograms, GIST, HOG, dense SIFT) for the task of video summarization. We put the input frames into the official pretrained GoogleNet model and fetch the output feature vectors of the fifth pooling layer; 2) We put the feature vectors into the LSTM pretrained in \cite{Zhang2016Video} and the outputs are the importances of the input frames we need. }
	\item Keyframe Relation: We compute the keyframe relation in keyframe selection stage. As mentioned before, we can select more than one frame in our keyframe selection and it is obvious that the selected keyframes with the same subtitle are semantic related. Note that, the result of the relation among keyframes is sparse because only the frames with the same subtitle can be regarded as semantic related. The result is shown in  Fig~\ref{fig_sim7}.
\end{itemize}

\textbf{Panel Allocation.} In a manga book, the number of panels on each page is not fixed, different pages have varied number of panels in order to make readers have a better reading experience. In our implementation, we take it as an optimization problem in a global way, and assign all the panels to manga pages together. 

Let  \{f$_1$,f$_2$,...,f$_N$\} denote the input keyframes where $N$ is the number of the input keyframes obtained from keyframe selection. Given the total number of pages $N$ in a manga book, we should generate the number of panels on each page subject to: 
\[\left\{ {\begin{array}{*{20}{c}}
	{n_i} \le {N_{\max }}\\
	{n_i} \ge {N_{\min }}\\
	{\sum\limits_{i = 1}^{M} {{n_i}}  = N}
	\end{array}} \right.\] 
Here, $n_i$ denotes the number of panels on the ith page. $N_{max}$ is the maximum number of panels on a page and $N_{min}$ is the minimum number of panels on a page. $M$ is the total number of pages, which is user-defined. There are two more constraints we should take into consideration. One is uniformity and the other is the relation among panels. Uniformity means that the panels with high importance rank should be assigned to different pages to avoid having boring pages, i.e., pages that only contain panels of low importance values. The relation among panels means that the panels with semantic relation should be assigned to the same page, and the readers can read it fluently. 

Considering of all the constraints above, we use genetic algorithm \cite{Harik1999Genetic} to solve panels allocation next. Let \{$n_1$,$n_2$,...,$n_{M}$\} denotes the number of panels on each page.In our implementation,we set the maximum number $N_{max}$ to 9. 

The most important part of  genetic algorithm is the valuation function. There are 5 parts in our valuation function. The first one is to calculate the absolute difference between each $n_i$ and $N_{max}$, ensuring that $n_i$ is no more than $N_{max}$. The second part is to calculate the absolute difference between each $n_i$ and $N_{min}$, so that $n_i$ is no less than $N_{min}$. The third part is to sum all $n_i$, and then calculate the absolute difference between the sum and N, which can ensure that the number of panels in the manga book is equal to the number of the input frames. The fourth part is to obtain the uniformity by calculating the standard deviation of the maximum importance $I_{i}$ on each page. The less the standard deviation is, the better the result is. The fifth part is to count the number of the panels with semantic relation but assigned to different pages. Taking the five parts as a whole, we calculated the weighted value of the five parts, which is shown in Equation (1) below:
\setlength{\arraycolsep}{0.0em}
\begin{eqnarray}
\min z &{}={}& {\alpha _1}\sum\limits_{i = 1}^{M} {\left| {{n_i} - {N_{\max }}} \right| + } {\alpha _2}\sum\limits_{i = 1}^{M} {\left| {{n_i} - {N_{\min }}} \right|}\nonumber\\ 
&&{+}\:{\alpha _3}\left| {\sum\limits_{i = 1}^{M} {{n_i} - N} } \right| + {\alpha _4}SD + {\alpha _5}R
\end{eqnarray}
\begin{eqnarray}
{SD} =std(I_1,I_2,...,I_M)
\end{eqnarray}
\setlength{\arraycolsep}{5pt}where ${SD}$ means the standard deviation and $R$ means semantic relation. Obviously, the first three parts have the most strict constraints, thus we assigned a greater coefficient to them. The last two parts have the most relaxed constraints, and we assigned a relatively smaller coefficient to them. Finally, the weighted value of the whole five parts is the valuation of the generation.

\textbf{Panel Layout.} Four main layout parameters have been computed by the aforementioned methods, then we input them into an existing panel layout method \cite{Cao2012Automatic}, which provides various layout styles learned from different manga series. It is noteworthy that their original method \cite{Cao2012Automatic} needs to manually pre-define all the parameters, that is unpractical to an automatic system.  Fig~\ref{fig_sim2}(b) shows the calculation of layout parameters and the corresponding output layout pages.

\subsection{Text Balloon Generation and Placement Module}
\begin{figure}
	\centering
	\subfigure[]{\includegraphics[width=1.25in]{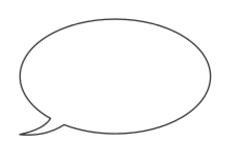}}
	\subfigure[]{\includegraphics[width=1.25in]{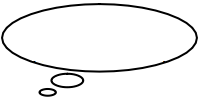}}
	\subfigure[]{\includegraphics[width=1.25in]{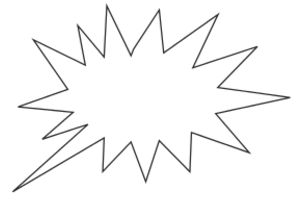}}
	\caption{Popular types of shapes of balloon.(a) rounded balloon. (b) thought balloon. (c) jagged contour balloon.}
	\label{fig_sim8}
\end{figure}

\textbf{Balloon Shape Selection.} It is very important to manga content expression that abundant of vivid balloon shape can be selected according to dialogues and emotion in various circumstances. However, existing system works generally only selects basic elliptical speech balloon shape for dialogues, which is sometimes insufficient to fully express a dialogue or an emotion. It is mentioned in \cite{Forceville2010Balloonics} that balloon shapes can help readers to capture information which does not have visual form such as sound. According to the statistic of \cite{Forceville2010Balloonics}, there are mainly 8 types of shapes of balloon. As shown in  Fig~\ref{fig_sim8}, the three most commonly used shapes are rounded balloon or rectangular straight balloon, thought balloon, and jagged contour balloon, sorted in descending order. Rounded balloon or rectangular straight balloon is usually used for common speech which makes it most likely to be used in manga creation. Thought balloon is always used to state the thoughts of characters in the panel without speaking it out. A jagged contour balloon is always used when there are verbal conflicts, such as screaming, etc. In our system, we consider the three common balloon shapes. 

\begin{figure*}
	\centering
	\includegraphics[width=5.5in]{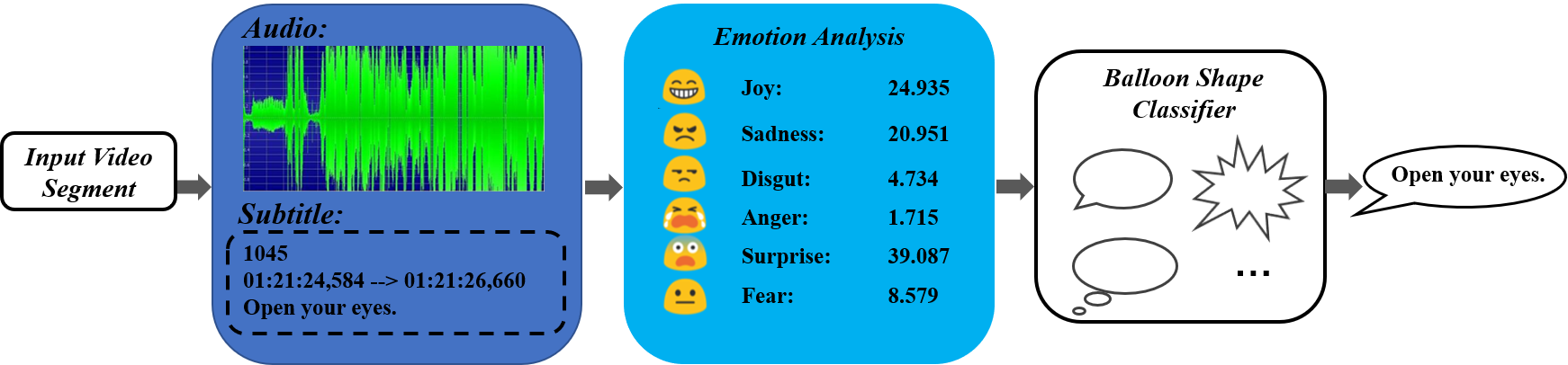}
	\caption{Emotion-aware Balloon Generation. Given an input video segment, we first get its subtitles and audio. Then, emotion analysis is performed to get the emotion of subtitles and audio. Finally, we feed the emotion values into a balloon shape classifier to predict the balloon shapes of the subtitles.}
	\label{fig_sim9}
\end{figure*}

Instead, we propose an emotion-aware balloon generation method, which can take advantage of videos' audio and subtitles containing emotional information to generate balloons with multiple kinds of shapes. As shown in  Fig~\ref{fig_sim9}, given a video segment, we determine the shape and size of the balloon for its subtitles. In particular, firstly, we obtain the emotion of subtitles and the corresponding audio by emotion analysis. Then, the emotion of the subtitle and the number of words will determine the size of the word in the balloon.  Finally, the shape of word balloon will be selected through a pre-trained classifier which takes as input the emotion of the audio and the subtitle for the panel and predicts the probabilities of different balloon types being selected. For example, if the emotion of the subtitle is plain, we tend to choose the common rounded balloon. If the emotion of the subtitle is strong, the jagged contour balloon should be chosen. 

To train our balloon shape classifier, we collect training data that contains three kinds of elements: audio emotion, subtitle emotion, and type of balloon. First of all, we collect some animes and corresponding manga books. For each balloon in a book, we record its type of shape (rounded, thought ,or jagged contour) and the text inside it. Then, using the text in the balloon, we can determine the time duration it belongs to in the anime. Meanwhile, we get the audio within this duration and put it into Mixed-Emotion \cite{Buitelaar2018Mixed} (an open-source emotion analysis tool). The outputs are valence and arousal of this period of sound. Valence is pleasantness (ranging from unhappy to happy) of a stimulus and arousal is the intensity of emotion (ranging from excited to calm). For the subtitles, we put it into an emotion recognizer and the output can represent the emotion of the subtitle. Here, we call the output emotion rank. Finally, we obtain the training data, which contains valence and arousal of the audio, emotion rank of the subtitle ,and the shape of the balloon. 

We select SVM (Support Vector Machine) as our classifier. We put the valence, arousal ,and emotion rank into the SVM. After several iterations, the SVM can learn the mapping relations between the input and the type of balloon shape. 
When in use, we put the valence and arousal of the audio and the emotion rank of the subtitle into the trained SVM classifier. Then, the output is the type of balloon shape we should use.

\begin{figure}
	\centering
	\includegraphics[width=4.5in]{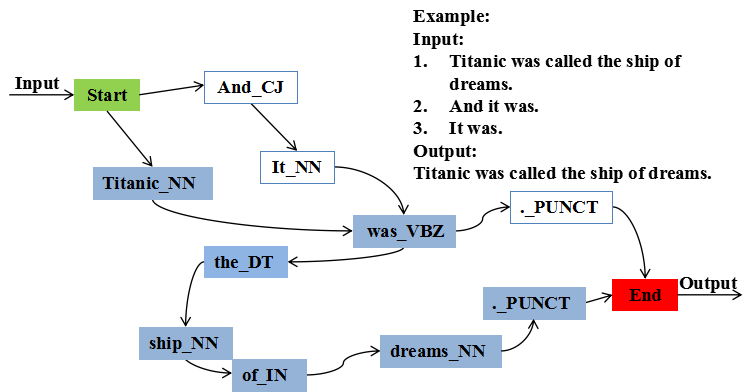}
	\caption{Example of the word graph with a possible compression path. For instance, the node `Titanic$\_$NN' in the word graph denotes that the word in the sentence is `Titanic' while `NN' indicates the part-of-speech of the word `Titanic' is a noun. The input is three sentences and finally output the combined result: Titanic was called the ship of dreams. The result is obtained by the path formed by the blue node.}
	\label{fig_sim10}
\end{figure}

\textbf{Text Summarization.} {It is possible that one keyframe corresponds to more than one sentence in our keyframe selection. If the sentence is too long for the word balloon, it will lead to an uncomfortable reading experience with repeated sentences and small font sizes.
	Hence, we use text summarization to merge multiple sentences into one compact sentence, and then render it on the generated balloon. To the best of our knowledge, we are the first to utilize text summarization in a comic generation system. }  As proposed by  \cite{Filippova2013Multi}, the basic idea of multi-sentence compression is to add a set of sentences (excluding punctuations) iteratively to the word graph. Each sentence has a pair of start and end nodes indicating the start and end of the sentence. The first sentence will be simply added to the graph. For the other sentences, a word is mapped into an existing node if they share the same part-of-speech and no word from the sentence has been previously mapped to the node. A new node can be created if there is no possible mapping.  Fig~\ref{fig_sim10} displays the word graph obtained from the set of sentences: 
\begin{itemize}
	\item Titanic was called the ship of dreams
	\item And it was.
	\item It was.
\end{itemize}
The node is in the form of a word and part-of-speech pair, separated by an underscore. For simplicity, the edge weight is omitted. 

To construct a word graph, words are mapped or created under three conditions:
\begin{itemize}
	\item Non-stop words without similar candidates in the graph or with an unambiguous mapping.
	\item Non-stop words with either several possible candidates in the graph or more than one occurrence in the same sentence.
	\item Stopwords.
\end{itemize}
Condition 2 and 3 will lead to ambiguous mapping. In this case, the preceding and following words in the sentence and the neighboring nodes in the graph will be the factors affecting the final mapping. Node with a larger overlap in neighboring words, or the one with more words mapped onto it, will be selected. After mapping/creation of nodes, the words in the sentence are connected with directed edges. New nodes or nodes which were not connected before will have an edge weight of one. Edge weights between previously connected nodes will increment by one. The mapped nodes also store a list of id of sentences that contain the word together with the corresponding position index in the sentences. Once all sentences are added to the graph, the word graph is completed and can be used to find the best compression of input sentences, in other words, the best-compressed sentence. It is done by finding the path among K shortest paths with lightest average edge weight. Before this, there are some restrictions on how the path weight is allocated. Individual edge weight $w_{ij}$ is given by Equation (4). 
\begin{equation}
\label{eqn_example}
association(i,j) = \frac{{freq(i) + freq(j)}}{{\sum\nolimits_{s \in S} {diff(s,i,j) - 1} }}
\end{equation}
\begin{equation}
\label{eqn_example}
w_{ij}= \frac{association(i,j)}{freq(i)\times freq(j)}
\end{equation}
where $freq(i)$ and $freq(j)$ are the numbers of words mapped to node i and j, respectively. Function $diff(s, i, j)$ refers to the distance between the positions of words $i$ and $j$ in sentence $s$, defined as Equation (5). Finally, K shortest path algorithm is implemented to find 50 shortest paths from start node to end node using Equation (4). Paths are rejected if they cannot satisfy the minimum number of words (suggested to be 8) or have no verb node. The remaining paths are ranked by a score that is calculated by normalizing the total path weight over its length. The path with the lightest average edge weight will be the best compression.
\begin{equation}
\label{eqn_example}
diff(s,i,j)\! =\! \left\{\!\!\! {\begin{array}{*{20}{c}}
	{pos(s,i) - pos(s,j),\!\!\!\!\!\!}\\
	{0,\!\!\!\!\!\!}
	\end{array}} \right.\begin{array}{*{20}{c}}
{\!pos(s,i)\! <\! pos(s,j)\!\!}\\
{\!otherwise\!\!}
\end{array}
\end{equation}

\begin{figure}
	\centering
	\includegraphics[width=4.5in]{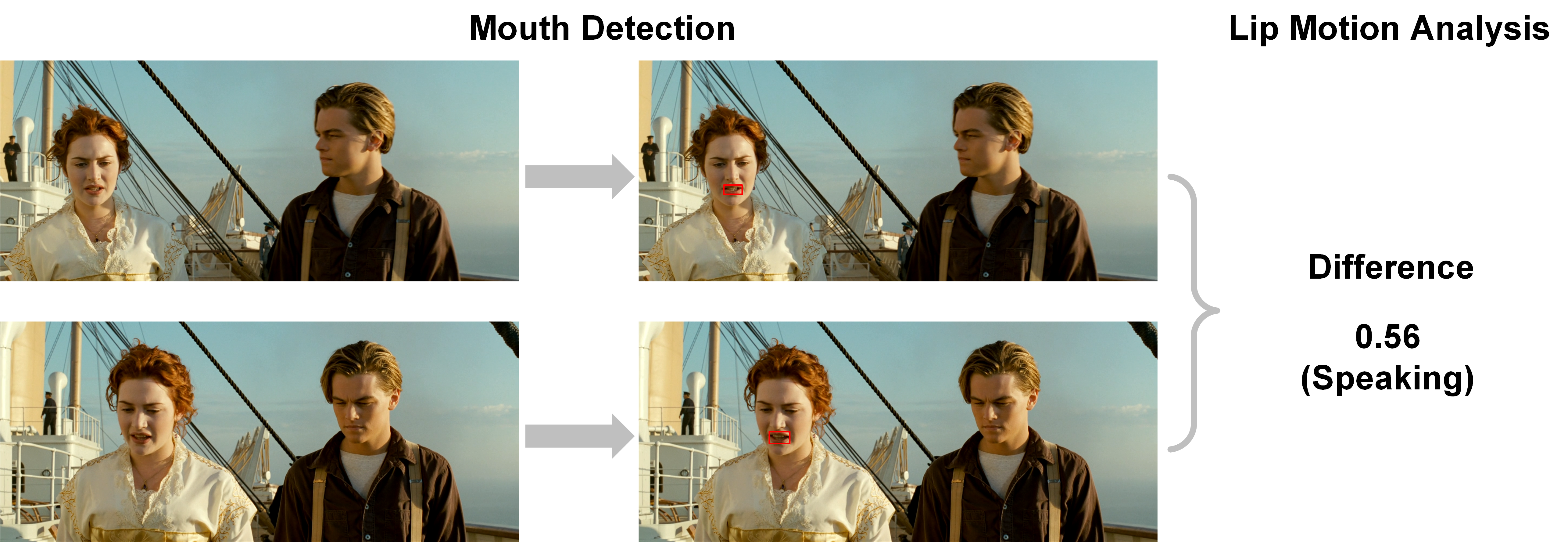}
	\caption{Pipeline of Speaker Detection.}
	\label{fig_sim11}
\end{figure}

\textbf{Balloon Placement.} Most existing work employed speaker detection and lip motion detection to get the location of the speaker in a frame, and then placed the balloon near the speaker it belongs to. Our approach is similar to theirs. As shown in  Fig~\ref{fig_sim11}, firstly, we detect the mouth of every character in a frame using a face-detector Python library called `Dlib', which is based on  Histogram of Oriented Gradients (HOG) to extract the feature points on a face. 68 feature points will be extracted from one face, of which 48th to 59th depict the outline of the mouth. Then, a lip motion analysis \cite{Kurlander1996Comic} is employed to compute the mean squared difference of the pixel values within the mouth region between two continuous frames. The difference is calculated over a search region around the mouth region in the current frame. Finally, a threshold is set to determine if a character is speaking or not. After we get the location of the speaker, we place the word balloon near the speaker and point the tail of the balloon to the speaker's mouth. 

\section{Experiments}
In this section, we will introduce the experimental results of our system. Firstly, we compared our results with a state-of-the-art comic generation system to present the superiority and aesthetic in our method. Then, we provide the results and evaluation of individual modules in our system, which include the efficiency of our system, accuracy and recall of importance rank, panel allocation, balloon generation ,and balloon placement. Thirdly, a subjective experiment is conducted to evaluate the quality of our results, as compared to a state-of-the-art comic generation system. 

\begin{table}[]
	\centering
	\caption{Information of Input Movies. }
	\label{my-label1}
	\begin{tabular}{c|ccc|c|ccc}
		\hline \hline
		Movie Name                     & Clip & Duration & Cost Time & Movie Name                     & Clip & Duration & Cost Time \\ \hline
		\multirow{4}{*}{Titanic}       & 1    & 3.25min  & 5.13min & 	\multirow{4}{*}{The Message}   & 1    & 2.15min  & 3.41min   \\
		& 2    & 2.58min  & 4.15min & & 2    & 2.56min  & 3.97min   \\
		& 3    & 4.26min  & 6.95min & & 3    & 5.15min  & 9.17min \\
		& 4    & 2.15min  & 5.56min & 	& 4    & 4.15min  & 6.51min\\ \hline
		\multirow{4}{*}{Friends}       & 1    & 1.73min  & 2.51min & \multirow{4}{*}{Up in the Air} & 1    & 5.61min  & 8.64min   \\
		& 2    & 2.72min  & 4.38min & 	& 2    & 4.22min  & 6.68min   \\
		& 3    & 2.15min  & 3.39min & & 3    & 4.57min  & 7.13min  \\
		& 4    & 4.79min  & 7.80min & & 4    & 3.28min  & 4.89min  \\ \hline \hline
	\end{tabular}
\end{table}

%

\subsection{Experiment Setting}
As shown in  \protect Table~\ref{my-label1}, the inputs come from 16 clips of 4 different movies, including Titanic, The Message, Friends, and Up in the Air. 	
The duration of input videos varies from 2 to 6 minutes and each clip from the four given movies has a subtitle to generate word balloons. For every clip, we record the consumed time to generate a manga book using our system and calculate the mean consumed time.

\begin{figure*}

	\subfigure[]{\includegraphics[width=1.30in,height=1.75in]{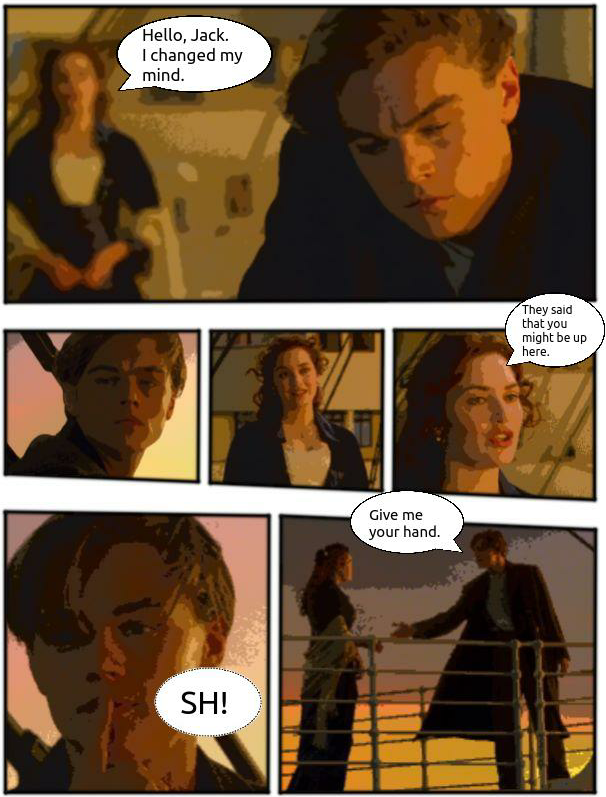}}
	\subfigure[]{\includegraphics[width=1.30in,height=1.75in]{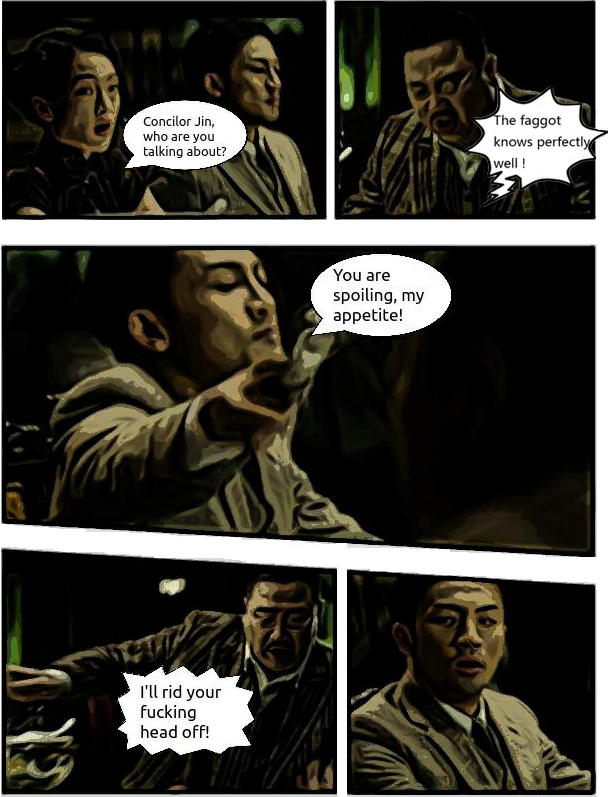}}
	\subfigure[]{\includegraphics[width=1.30in,height=1.75in]{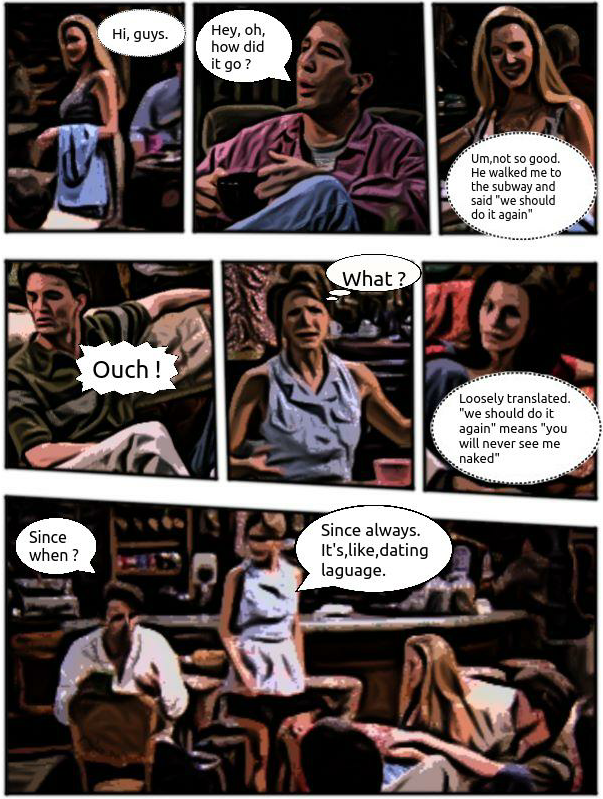}}
	\subfigure[]{\includegraphics[width=1.30in,height=1.75in]{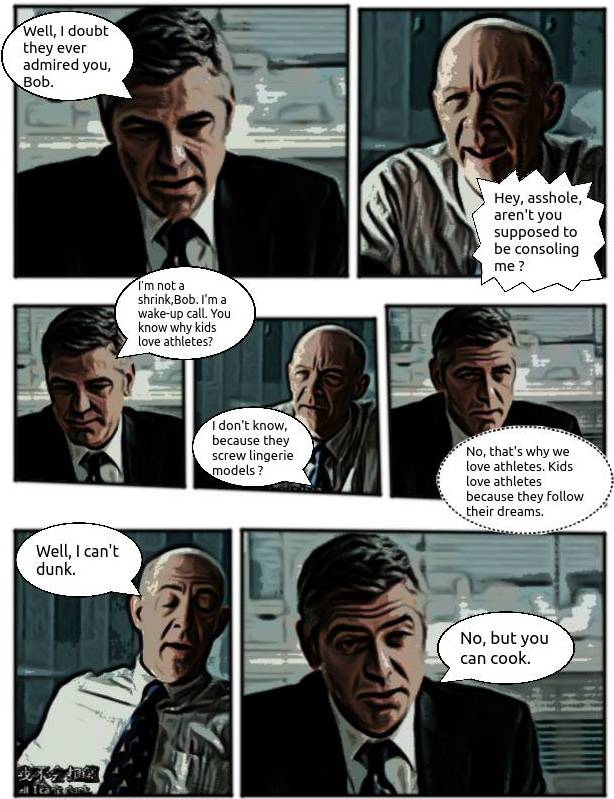}}
	\subfigure[]{\includegraphics[width=1.30in,height=1.75in]{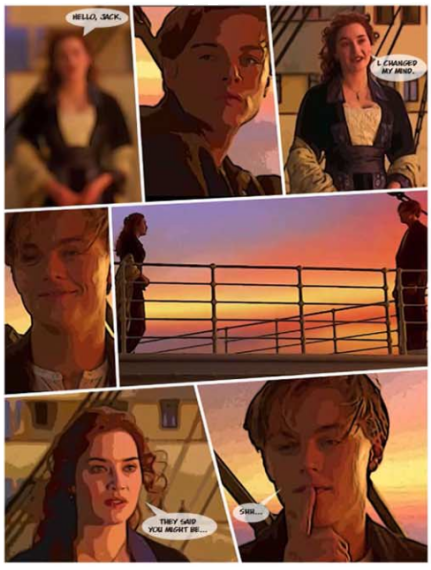}}
	\subfigure[]{\includegraphics[width=1.30in,height=1.75in]{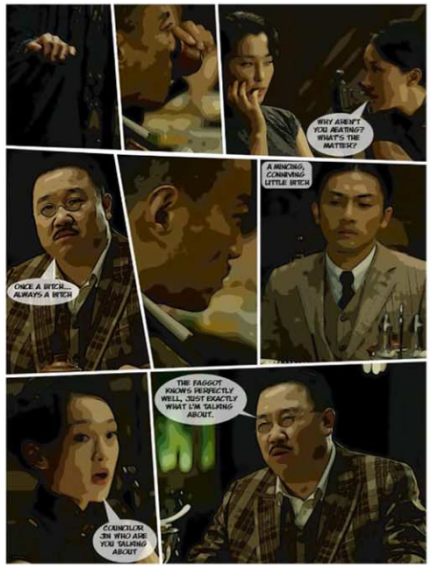}}
	\subfigure[]{\includegraphics[width=1.30in,height=1.75in]{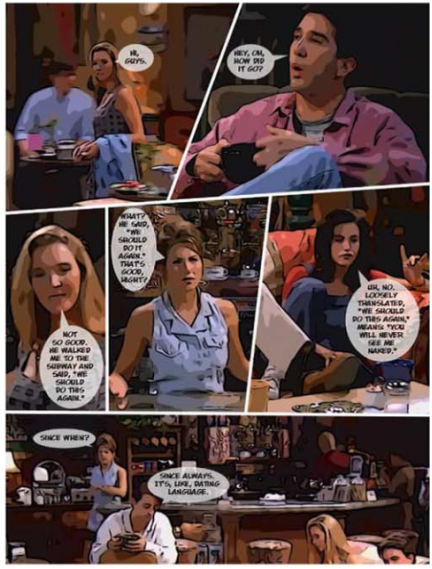}}
	\subfigure[]{\includegraphics[width=1.30in,height=1.75in]{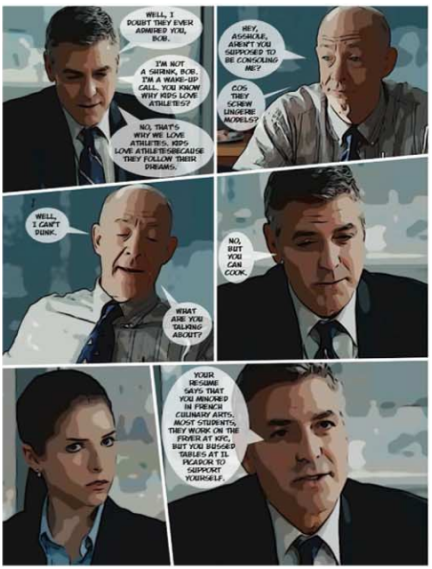}}
	\caption{Comparisons of our method with Content-Aware Video2Comics \cite{Jing2015Content}. (a)-(d) are our results. (e)-(h) are the results of \cite{Jing2015Content}. (a) and (e): Titanic (1997) (20th Century Fox, Paramount Pictures and Lightstorm Entertainment). (b) and (f): The Message (Huayi Brothers). (c) and (g): Friends [Bright/Kauffman/Crane Productions, Warner Bros. Television, NBC and Warner Bros. Television Distribution (worldwide)]. (d) and (h): Up in the Air (DW Studios, The Montecito Picture Company, Rickshaw Productions and Paramount Pictures).}
	\label{fig_sim12}
\end{figure*}

\subsection{Visual Comparison with Prior Work}
In this part, we will show some end results of our system and compare them with the results of some existing works \cite{Jing2015Content}. As shown in  Fig~\ref{fig_sim12}, our work is superior to the other comparison methods in three aspects. Firstly, our system can generate more abundant balloon shapes for word balloon, instead the existing methods only use simplex elliptical word balloon. Secondly, we employ text summarization to merge some related subtitles so that we can ensure the sentence in a word balloon is not too long. Thirdly, we provide fully automatic multi-page layouts by obtaining four important parameters automatically. And the result of our layout behaves reasonable and abundant. 

\subsection{Evaluation of Individual Modules}


\begin{table}[]
	\centering
	\caption{{{Accuracy of our importance rank detection method. Our learning strategy achieves accurate enough results for our importance ranking task. }}}
	\label{my-label2}
	\begin{tabular}{c|ccc|c|ccc}
		\hline
		\hline
		\small{Movie Name}      & \small{Clip} & \footnotesize{Number of Frames} & \small{Accuracy}
		& \small{Movie Name}                     & \small{Clip} & \footnotesize{Number of Frames} & \small{Accuracy}  \\ \hline
		\multirow{4}{*}{\small{Titanic}}       & 1    & 393              & 0.796  & 	\multirow{4}{*}{\small{The Message}}   & 1    & 258              & 0.751   \\
		& 2    & 305              & 0.768 & & 2    & 308              & 0.838     \\
		& 3    & 511              & 0.714 & & 3    & 622              & 0.741   \\
		& 4    & 407              & 0.758  & & 4    & 491              & 0.750 \\ \hline
		\multirow{4}{*}{\small{Friends}}       & 1    & 210              & 0.742 & \multirow{4}{*}{\small{Up in the Air}} & 1    & 675              & 0.791    \\
		& 2    & 326              & 0.802 & & 2    & 334              & 0.783   \\
		& 3    & 258              & 0.763 & & 3    & 548              & 0.762   \\
		& 4    & 571              & 0.745 & & 4    & 397              & 0.812  \\ \hline\hline
	\end{tabular}
\end{table}

\textbf{Importance Rank.} Importance rank is used to allocate importance to panels on a page. In our implementation, we utilize the neural network proposed in \cite{Zhang2016Video} to learn and predict the panel importance. In order to validate the accuracy of this learning approach, we also compare the results with the ground truth that are manually labeled by human.  \protect Table~\ref{my-label2} shows that our designed learning strategy is accurate enough for this purpose. As for the calculation of accuracy, for each video clip, we asked ten people to mark the importance of the panels ranking from 1 to 4. The manually obtained importance is the ground truth, and then, we compare it with our results to compute the accuracy of our method.


\begin{table}[]
	\centering
	\caption{{{Accuracy of panel allocation method. We compare our results with manually allocated panels and our method achieves reasonable performance. }}}
	\label{my-label3}
	\begin{tabular}{c|cc|c|cc}
		\hline\hline
		Movie Name                     & Clip & Accuracy & Movie Name                     & Clip & Accuracy \\ \hline
		\multirow{4}{*}{Titanic}       & 1    & 0.713 & \multirow{4}{*}{The Message}   & 1    & 0.625    \\
		& 2    & 0.654 & & 2    & 0.580    \\
		& 3    & 0.697 & &  3    & 0.698   \\
		& 4    & 0.665  & & 4    & 0.642 \\ \hline
		\multirow{4}{*}{Friends}       & 1    & 0.640  & \multirow{4}{*}{Up in the Air} & 1    & 0.715  \\
		& 2    & 0.691  & & 2    & 0.599   \\
		& 3    & 0.593  & & 3    & 0.625  \\
		& 4    & 0.585  & & 4    & 0.655  \\ \hline\hline
	\end{tabular}
\end{table}


\textbf{Panel Allocation.} Panel Allocation is used to allocate keyframes to different pages. To evaluate all these constraints, we employ Equation (1). In our implementation, the coefficient $\alpha$$_1$-$\alpha$$_5$ are set to 3,3,2,1,1 respectively which can get the best results. To validate the results, we asked ten people to allocate panels manually for each video clip. Each allocation can be represented by a numerical sequence, and we can compare our results with manual obtained allocation numerical sequence to compute the accuracy. The result is shown in  \protect Table~\ref{my-label3}. Our Panel Allocation Method is accurate and reasonable enough for this module.

\begin{table}[]
	\centering
	\caption{{{Accuracy and recall of our panel shape classifier. Among three different kinds of classifiers, support vector machine (SVM) works the best with the highest accuracy and medium recall. }}}
	\label{my-label4}
	\begin{tabular}{c|cc}
		\hline\hline
		Model   & Accuracy & Recall \\ \hline
		SVM          & 0.994    & 0.835  \\
		DecisionTree & 0.959    & 0.857  \\
		RandonForest & 0.969    & 0.796  \\ \hline\hline
	\end{tabular}
\end{table}

\textbf{Balloon Shape Selection.} Firstly, for each subtitle, we select a balloon shape for it manually. The ground truth we get in this step is used to train a classifier that will be used for balloon shape selection. Then we get the emotion rank of the subtitles and the valence and arousal score of their corresponding audio with an existing tool "Mixed Emotion''. Finally, we put the data we get before into a classifier for training. In our implementation, we employ three different kinds of classifiers: support vector machine (SVM), DecisionTree, and RandomForest. The accuracy and recall are shown in  \protect Table~\ref{my-label4} and it shows that SVM works best because the accuracy of SVM is highest and the recall of it is medium.

\begin{figure*}
	\centering 
	\subfigure[]{\label{fig:subfig2:a}
		\includegraphics[width=1.2in,height=0.72in]{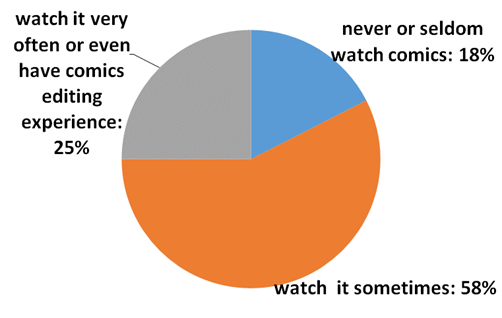}}
	\subfigure[]{\label{fig:subfig2:b}
		\includegraphics[width=1.2in,height=0.72in]{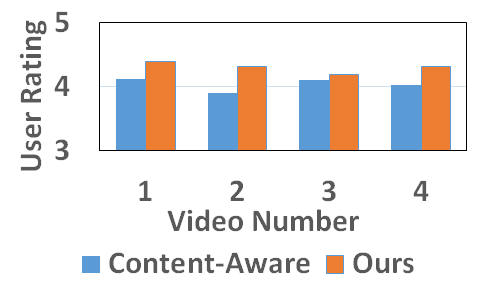}}
	\subfigure[]{\label{fig:subfig2:c}
		\includegraphics[width=1.2in,height=0.72in]{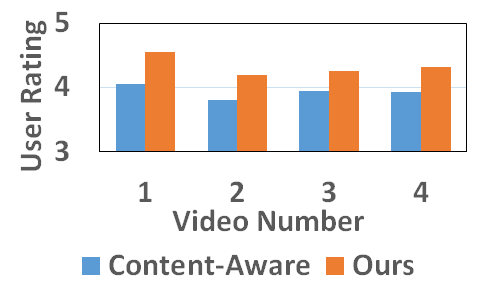}}
	\subfigure[]{\label{fig:subfig2:d}
		\includegraphics[width=1.2in,height=0.72in]{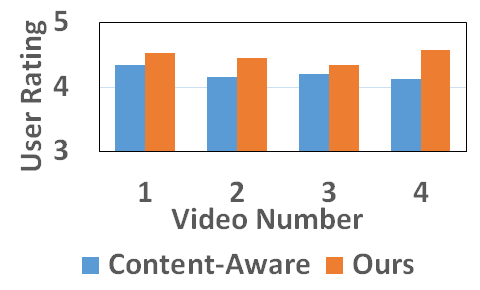}}

	\caption{{(a) refers to the information about users' familiarity with comics. (b), (c) and (d) are the average score of each question for the users who never or seldom watch comics, who watch it sometimes and who watch it very often or even have comics editing experiences, respectively.  }}
	\label{fig:subfig2}
\end{figure*}

{{\subsection{User Study} 
		
		
		We also conduct a user study to further evaluate the effectiveness of our system, and four questions are included in our questionnaire. In our experiments, we 
		recruit 40 participants via Amazon Mechanical Turk to compare our results with those by \cite{Jing2015Content} and evaluate different aspects of the results. The participants are first asked to watch a video and then read the comics generated by either our method or \cite{Jing2015Content}, then they answer several questions regarding different aspects of generated comics by rating them using a scale from 1 to 5, with 1 being the worst and 5 being the best. 
		The videos and the corresponding comics are presented in a random order to avoid subjective bias.   The aforementioned questions are listed as follows:
		
		\begin{figure*}
			\centering 
			
			\subfigure[Overall]{\label{fig:subfig:f}
				\includegraphics[width=1.0in,height=0.61in]{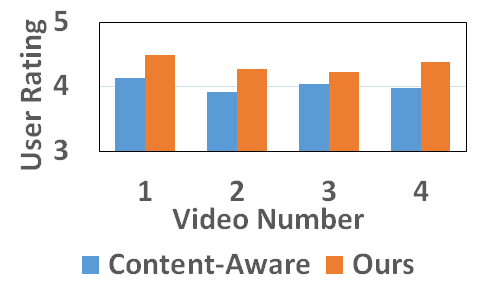}}
			\subfigure[Balloon quality (Q1) ]{\label{fig:subfig:g}
				\includegraphics[width=1.0in,height=0.61in]{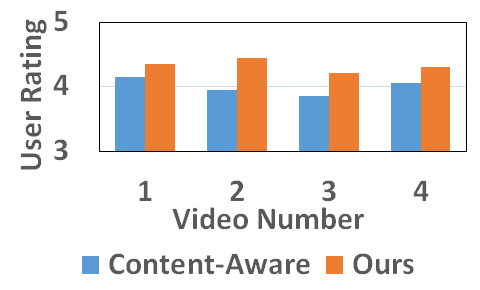}}
			\subfigure[Layout quality (Q2)]{\label{fig:subfig:h}
				\includegraphics[width=1.0in,height=0.61in]{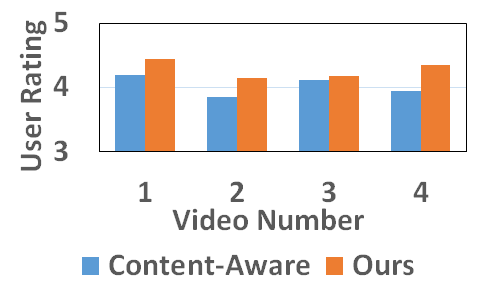}}
			\subfigure[Storytelling (Q3)]{\label{fig:subfig:i}
				\includegraphics[width=1.0in,height=0.61in]{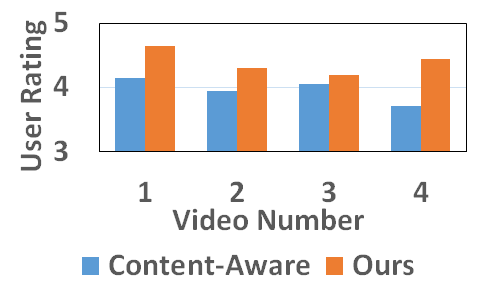}}
			\subfigure[Reading experience (Q4)]{\label{fig:subfig:j}
				\includegraphics[width=1.0in,height=0.61in]{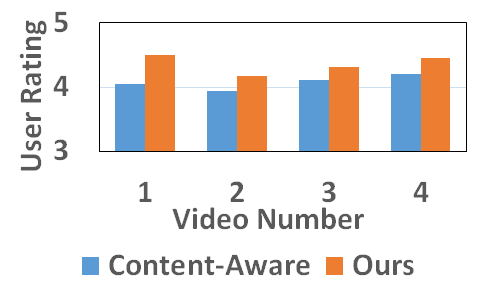}}
			\caption{{Results of User Study. The vertical axis is the average rating score of each question. The horizontal axis denotes the video number,1,2,3,4 stand for the films  Titanic (1997) (20th Century Fox, Paramount Pictures and Lightstorm Entertainment), The Message (Huayi Brothers), Friends [Bright/Kauffman/Crane Productions, Warner Bros. Television, NBC and Warner Bros. Television Distribution (worldwide)] and Up in the Air (DW Studios, The Montecito Picture Company, Rickshaw Productions and Paramount Pictures). According to an unpaired t-test, all the preferences are statistically significant (p $ < $ 0.05).}}
			\label{fig:subfig}
		\end{figure*}

		\begin{description}
			\item[Q1:] To what degree do you think the quality of balloons (in style and position)?
			\item[Q2:] To what degree do you think the visual quality of layouts?
			\item[Q3:] To what degree do you think the ability of comics in expressing the video contents?
			\item[Q4:] To what degree do you think the ability of comics in providing you with engaging reading experience? 
		\end{description}

		\textbf{Result.} As shown in  Fig~\ref{fig:subfig} (a)-(j), for every question and movie, our system performs better than the method \cite{Jing2015Content} (content-aware) no matter whether participants have watched the videos before. According to the readers' feedback (the data can be obtained in the supplementary file), emotion-adaptive varying balloon types and font sizes make our results look visually richer and can better express the stories, while \cite{Jing2015Content} only use one type of word balloon. Also, our layouts are visually diverse and closely resembles real manga styles. \\
		We try to further analyze the results of our subjective evaluations statistically. Using an unpaired t-test, we find that, for each of Q1-Q4, there is a statistically significant difference in subjects choosing our method over the method \cite{Jing2015Content} (all p-value $ < $ 0.05). This is expected, as our method uses multi-page semantically rich panel layouts and emotion-aware word balloons together, and thus has the ability to generate a higher quality of balloons (Q1), design more attractive layouts (Q2), present better visual contents (Q3), and all together make our manga presentation more natural and delightful (Q4).
}}\\
{We also analyze the responses of different types of users based on their familiarity with comics. As shown by Fig~\ref{fig:subfig2:a}, we analyze our results of three types of users: 1) those who know nothing about comics or seldom read comics, 2) those who sometimes read comics and 3) those who often read comics or even have comics editing experience. The average rating scores are summarized in Fig~\ref{fig:subfig2}(b)-(d). Our methods performs better in each group of users. The unpaired t-test demonstrates that, there are statistically significant differences (all p-value $<$ 0.05) for each group. 	}

\section{Conclusion}
In this paper, we have presented a system that is capable of generating high-quality comics without any user inputs. At the core of our system is a multi-page layout framework that jointly organizes multiple pages in visually rich layouts using rich semantics from videos, and an emotion-aware balloon generation method that can create a variety of balloon shapes and font sizes according to emotion contained in subtitles and audio. Our experiments demonstrate that our system can synthesize more expressive and engaging comics, in comparison to a state-of-the-art comic generation system. 
Although our system has been shown to achieve promising results, it is still subject to several limitations. For example, the keyframe selection is not accurate enough. In some cases, the selected keyframes are similar to each other,  which would certainly introduce redundancy into generated comics. It would be interesting to develop an end-to-end neural network for keyframe selection, which is left for future work. Furthermore, it is time-consuming to generate comics from a video without any subtitles. That is because speech recognition, which is required for subtitle extraction, is often prone to errors, and thus many manual efforts are needed to refine the recognition results. We believe that, with the advent of more sophisticated speech recognition technique, this problem can be alleviated greatly. What's more, inspired by many existing methods \cite{ravi2018show, chen2019neural, li2019storygan} which can generate image sequences given a story with multiple sentences, it is possible to produce comic books from textual stories and we are interested to extend our method to leverage textual information to help generating manga.

\bibliographystyle{ACM-Reference-Format}
\bibliography{sample-bibliography}

\end{document}